\documentclass[10pt,a4paper,conference]{IEEEtran}

\usepackage{amsmath,amsfonts,amssymb}
\usepackage{xcolor}
\usepackage{multirow}

\definecolor{modifica}{rgb}{1.0, 0.75, 0.0}

\ifCLASSINFOpdf
  \usepackage[pdftex]{graphicx}
  \graphicspath{{../pdf/}{../jpeg/}{../png}}
  \DeclareGraphicsExtensions{.pdf,.jpeg,.png}
\else
\fi

\begin{document}
% Titles are generally capitalized except for words such as a, an, and, as, at, but, by, for, in, nor, of, on, or, the, to and up, which are usually not capitalized unless they are the first or last word of the title.
\title{Subspace Clustering for Action Recognition with Covariance Representations and Temporal Pruning}

% use a multiple column layout for up to three different affiliations
\author{
\IEEEauthorblockN{Giancarlo Paoletti, Jacopo Cavazza, Cigdem Beyan and Alessio Del Bue}
\IEEEauthorblockA{Pattern Analysis and Computer Vision, Istituto Italiano di Tecnologia, via Enrico Melen 83, 16152 Genova, Italy\\
%\texttt{name.surname@iit.it}
\{\texttt{giancarlo.paoletti,jacopo.cavazza,cigdem.beyan,alessio.delbue}\}\texttt{@iit.it}}}

% conference papers do not typically use \thanks and this command is locked out in conference mode.
% If really needed, such as for the acknowledgment of grants, issue a \IEEEoverridecommandlockouts after \documentclass


\maketitle

% As a general rule, do not put math, special symbols or citations in the abstract
\begin{abstract}
This paper tackles the problem of human action recognition, defined as classifying which action is displayed in a trimmed sequence, from skeletal data.
% and focuses on skeletal data, which are superior to traditional RGB videos since better handling visual ambiguities while preserving privacy. 
Albeit state-of-the-art approaches designed for this application are all supervised, in this paper we pursue a more challenging direction: Solving the problem with \emph{unsupervised learning}. To this end, we propose a novel subspace clustering method, which exploits covariance matrix to enhance the action's discriminability and a timestamp pruning approach that allow us to better handle the temporal dimension of the data.
%two alternative, yet, complementary strategies to handle the temporal dimensions of the data: covariance representations and timeframes selection. \textcolor{red}{[FIXME: give a quick explanation why these two are fundamental.]}  %J: Done.
Through a broad experimental validation, we show that our computational pipeline surpasses existing unsupervised approaches but also can result in favorable performances as compared to supervised methods. % improves in performance not just clustering-based methods but can even favorably scores with respect to supervised approaches.
\end{abstract}

\IEEEpeerreviewmaketitle

%%%%%%%%%%%%%%%%%%%%%%%%%%%%%%%%%%%%%%%%%%%%%%%%%%
\section{Introduction}\label{sec:intro}
%\textcolor{red}{Say something about the pipeline (COV).} 

Human Action Recognition (HAR) plays a crucial role in 
%pattern recognition 
computer vision since related to a broad spectrum of artificial intelligence %computer vision 
applications (such as video surveillance, human-machine interaction or self-driving cars to name a few \cite{survey_Poppe}). Given a trimmed sequence, in which a single action or activity is assumed to be present, %encoded in it,
the final goal of HAR is to correctly classifying it. Although significant progresses have been made in the last years, accurate action recognition in videos is still a challenging task because of the complexity of the visual data e.g., due to varying camera viewpoints, occlusions and abrupt changes in lighting conditions.     % due to the many visual ambiguities such as viewpoint, occlusions or varying lighting conditions. 
As an all-in-one solution to these problems, skeleton-based HAR is surely the paradigm to embrace, considering also its beneficial characteristics of being privacy-preserving. In skeleton-based HAR, action/activity sequences are represented through the multi-dimensional time series of joints, located at the intersection of skeletal bones, whose position is tracked in time typically through either motion capture systems or depth sensors. %\textcolor{red}{[FIXME: here a figure summarising Figure 1 and showing the input data and the problem would be very useful]}.

Recently, skeleton-based HAR has undergone to the same paradigm shift which was registered in other fields of pattern recognition: Hand-crafted data encodings fed into engineered classifiers have been replaced by data-driven feature representation with an end-to-end classification pipeline \cite{NIPS2012_4824}. Yet, both paradigms leverage a fully \emph{supervised} learning approach to accomplish the task. Each sequence is in fact assumed to be (manually) annotated by the action/activity it involves. Other than being time-consuming and prone to human errors, sequence annotations compromise the scalability to the big data regime. As an alternative, unsupervised approaches seem attractive since they offer an advantage regarding computational and methodological burden, as well as providing an interesting application towards more novel real-life scenarios. %\textcolor{red}{[FIXME: you cut too short this sentence, be very convincing why unsupervised methods are better]}. 

In this work, we consider \emph{subspace clustering} %\textcolor{orange}{(specifically the implementation of \cite{vidal2011subspace} as reference)} \textcolor{red}{[FIXME: Rene is not the first using subspace clustering, of course he is the most famous]} %J: I will remove the citation. I do not see any other paper which is more protypical and appropriate to cite
to tackle HAR in a fully unsupervised paradigm. Subspace clustering was first introduced in Computer Vision to segment dynamic moving objects \cite{gear1998multibody,costeira1998multibody} and it postulates that high-dimensional data (here, skeletal joints) can be represented as a union of subspaces, each of them having a much lower dimensionality (i.e. low-rank) and simpler geometrical structure. Each subspace usually corresponds to a class (here, to an action or an activity). The key idea in subspace clustering is to learn encodings that are then used to construct an affinity matrix from which the data can be clustered together according to the modelled (dis)-similarities between samples \cite{vidal2011subspace}. Although, this is usually achieved through a self-expressive model in which each data point is expressed as a linear combination of the remaining ones, additional constraints, such as sparsity, were also adopted \cite{elhamifar2013sparse}.
% \textcolor{red}{[FIXME: this is one of the ways to estimate this encoding, but not the only one...be more balanced when doing these statements...]}.

Despite the fact that subspace clustering has become a powerful technique for problems such as face clustering or digit recognition, its applicability to the problems like skeleton-based HAR was only explored by a limited number of works \cite{zhang2012improving,li2015temporal,clopton2017temporal}. This is due to many operative limitations including how to handle the temporal dimensions, the inherent noise present in the skeletal data and the related computational issues. %related on how to handle the temporal dimensions of the data encoding the actions to be recognized. \textcolor{orange}{It is worth to mention also the noisy nature of many HAR datasets, caused often by the nature of the capturing devices, or the hindered computational efficiency of available methods which can lead to computational OOMs etc.} \textcolor{red}{[FIXME: expand here with a couple of sentences about standard shortcomings: computational explosion, noisy data, etc...].}

In this paper, we propose two alternative computational strategies to help and support subspace clustering methods in handling the temporal dimensions of action sequences. On the one hand, we encode the raw skeletal trajectories using a covariance representation, which has been shown to be effective for the solving HAR problems \cite{cavazza2019scalable}.
%. Additionally, covariance has the additional desirable property of being capable of encode action instances even if they have variable temporal duration. 
Additionally, we propose a computational strategy to prune the instantaneous body poses -- termed \emph{timestamps} hereafter -- whose temporal aggregation produces an action sequence. As the result of temporal pruning, we are able to select the most representative timestamps, which are exploited to compress the original action sequence to a fixed duration. Consequently, this \emph{temporal pruning} can be adopted as a successful pre-processing step to accommodate for the usage of a subspace clustering method for HAR. %\textcolor{red}{[FIXME: there is a contrast in what you say...covariance can accomodate arbitrary long sequences but then temporal pruning get you with a fixed time sequence...you might mislead that covariance then is still not so interesting. Please rephrase.]} %J: I removed the sentence which in my opinion originated the issue. I think it is ok now.
%\textcolor{blue}{both cov and pruning get a fixed temporal length, we want to demonstrate that if you treat skeleton dataset as a collection of point per-se (like SSC does and all the self-expressiveness data), you should use cov because temporal info gets encoded well and can give you an acc boost, w.r.t. vanilla skeleton data. Because you encode temporal data in a higher level representation. Instead if you want to treat temporal data without representations, you must prune to a fixed length, on how you do it it depends on the different strategies we performed in the paper, or you can replicate lower timestamps of a dataset to get all length(max dataset timestamp). Long-story-short if you use dictionaries and temporal laplacian regularizers (like used in TSC) this temporal pruning gives you top results, even comparable to supervised accs.}.

%in which the main experimental validation was aimed at demonstrating that a few methods for subspace clustering can outperform other clustering-based approaches on a limited set of motion capture datasets. Differently, in this paper, we %propose the first extended experimental validation in which we ablated on several subspace clustering methods, in order to 
%\textcolor{red}{To write again this sentence:With a comprehensive experimental analyses, we provide a validation to dissect the alternative of covariance representation and temporal pruning when alternatively paired with subspace clustering.} 
Through a comprehensive experimental analysis, we validate the impact on HAR of covariance representations and temporal pruning. 
Eventually, we also demonstrate their degree of complementary to the extent that the performance of a fully unsupervised recognition pipeline can be enhanced. Surprisingly, the overall performance of the proposed unsupervised approaches can almost fill the gap with state of the art supervised methods. %between the one achieved through approaches and the ones obtained with fully supervised regimes. %(FIXME: sentence too long and verbose, rephrase).  

Overall, we deem that our experimental findings would help practitioners in re-thinking the way HAR is approached, raising the attention in the desirable shift towards more agile unsupervised learning frameworks.

%%%%%%%%%%%%%%%%%%%%%%%%%%%%%%%%%%%%%%%%%%%%%%%%%%
\section{Related Work}\label{sec:relatedWork}
In this Section, we have reviewed the action recognition methods relying on covariance representation, various subspace clustering algorithms as well as the state-of-the-art supervised approaches for skeleton-based HAR. %\textcolor{red}{G: add here TSC part.}
\newline
\textbf{Subspace clustering.}
Subspace clustering has been a popular computational framework in the machine learning community as well as the computer vision and image processing communities (e.g., image representation and compression \cite{Hong2006}, image segmentation \cite{Yang2008}, motion segmentation \cite{Fan2006}).
It aims at finding subspaces each ﬁtting a group of data points and then performing clustering based on these subspaces \cite{vidal2011subspace}.

There has been a lot of work presenting many different subspace clustering methods. Most of the subspace clustering methods learns an afﬁnity matrix and then apply spectral clustering, e.g., low-rank representation \cite{Liu2013,Lu2019}. Self-representation based subspace clustering methods reconstructs a sample from a linear combination of other samples \cite{elhamifar2013sparse,Liu2013,Hu2014,Lu2012} and they have proven their effectiveness for high-dimensional data.
Sparse subspace clustering integrates l1-norm regularization, which mostly results in improvements in the clustering performances \cite{elhamifar2013sparse}. The temporal Laplacian regularization was proposed in \cite{li2015temporal} and also adopted in other works e.g., \cite{clopton2017temporal} to better model kinematic data for the sake of action detection and segmentation.

Most existing subspace clustering methods relies on handcrafted representations. Instead, more powerful representations can be learned through deep learning, which effectively cluster data samples from non-linear subspaces \cite{ji2017deep}. Deep subspace clustering methods apply embedding and clustering jointly, typically with an autoencoder network e.g., in \cite{ji2017deep,Yang2019}. This results in an optimal embedding subspace for clustering, which is more effective compared to conventional clustering methods. %Deep adversarial subspace clustering methods, on the other hand, learns sample representations by deep learning for subspace clustering but specifically introduces adversarial learning such that a subspace clustering generator and a quality-verifying discriminator are learned against each other 
%Deep adversarial subspace clustering methods, on the other hand, learn more effective sample representations using deep learning while exploiting adversarial learning to supervise and, thus, progressively improve the performance of subspace clustering. This is done by using a subspace clustering generator and a quality-verifying discriminator which are adversarially learned against each other \cite{Zhou2018}.In that way, they learn more effective sample representations using deep learning while exploiting adversarial learning to supervise and, thus, progressively improve the performance of subspace clustering.
Deep adversarial subspace clustering methods, on the other hand, learn more effective sample representations using deep learning while exploiting adversarial learning to supervise and, thus, progressively improve the performance of subspace clustering. This is done by using a subspace clustering generator and a quality-verifying discriminator which are adversarially learned against each other.
%, which performs representation learning and subspace clustering. Thus, clustering performance can be evaluated in addition to the self-representation loss \cite{Zhou2018}.  %\textcolor{red}{G: add here TSC part.}
%a novel deep adversarial subspace clustering (DASC) model, which learns more favorable sample representations by deep learning for sub-space clustering, and more importantly introduces adversarial learning to supervise sample representation learning and subspace clustering. Specifically, DASC consists of a subspace clustering generator and a quality-verifying discriminator, which learn against each other.
\newline
\textbf{Covariance encoding for HAR.} The idea of encoding 3D-skeleton dynamics within a single hand-crafted kernel representation has been proposed often in HAR. For instances, it has been shown that Hankel matrices can efficiently model action dynamics when used in tandem with a Hidden Markov Model \cite{Camps:ACCV14} or a Riemannian nearest neighbours with class-prototypes \cite{Camps:CVPR16}. Lie group \cite{Vemulapalli:CVPR14} and associated Lie algebra \cite{Vemulapalli:CVPR16} can be effective in modelling human actions and activities by means of roto-translations. Likewise, generic deforming bodies can be efficiently modelled over variations of Stiefel manifolds \cite{delbue2011bilinear}. Surely, within the class of kernel representations, a major role is played by a specific symmetric and positive definite (SPD) operator: Covariance matrices (COV). Originally envisaged for image classification and detection \cite{TPM:ECCV06}, COV is an effective representation for skeleton-based HAR since capable of modelling second-order statistics. It was used in tandem of a variety of classification pipelines, such as a temporal pyramid \cite{egizi} or max-margin approaches \cite{Wang:ICCV15,ECCV16}. Formal studies have tried to enhance the capability of such operators in modelling non-linear correlations among the data \cite{Harandi:CVPR14,Cavazza:ICPR16}. Kernel approximation was recently investigated in order to speed up the computational pipeline and ensure scalability towards the big data regime \cite{Cavazza2019}.

Even though prior work focused on the effectiveness of covariance representations applied to supervised learning pipelines, we instead demonstrate its capabilities for unsupervised learning.
\newline
\textbf{State-of-the-art supervised approaches for skeleton-based HAR.} The current mainstream paradigm in skeleton-based HAR is the possibility of learning a feature representation from the data itself, in tandem with the final action classifier. As one of the seminal works in this direction, a hierarchy of bidirectional recurrent neural networks is used by \cite{Du:CVPR15} to represent in a bottom-up fashion all the structural relationships between body parts (torso, legs, arms) in the human skeleton. Long-Short Term Memory (LSTM) models can be proficiently applied to 3D action recognition \cite{Shahroudy:CVPR16,Liu:ECCV16}. Throughout the years, LSTM networks have been modified to better accommodate for the task: for instance, by applying a novel mixed-norm regularization term and dropout \cite{Zhu:AAAI16} or recurring to attention mechanisms \cite{Liu:CVPR17}. Alternatively, joint trajectories are casted into colored images by producing the so-called distance maps \cite{JCNN1,JCNN2,Ke:CVPR17}. By means of them, usual convolutional neural networks such as AlexNet, despite originally proposed for image classification, can be adapted to HAR \cite{JCNN1,JCNN2}.
Surely, the most active and recent direction of research leverages the possibility of encoding the whole human skeleton as a graph, furthermore processing it through a graph-convolutional neural network \cite{shi2019two,wu2019spatial}.

All such approaches can fully exploit the benefits of an end-to-end and data-driven training since relying on a fully supervised regime in which the sequences to be classified are annotated. Differently, in this paper we pursue the more challenging direction of adopting an unsupervised strategy, relying on subspace clustering. Similarly to what done by \cite{su2019predict} for auto-encoders and \cite{zheng2018unsupervised} for generative adversarial networks, the goal of this paper is to propose new computational architectures and evaluate their effectiveness in comparison with supervised learning paradigms.

%%%%%%%%%%%%%%%%%%%%%%%%%%%%%%%%%%%%%%%%%%%%%%%%%%
\section{Methodology}\label{sec:method}

\begin{figure*}[t!]\centering
\includegraphics[width=\textwidth]{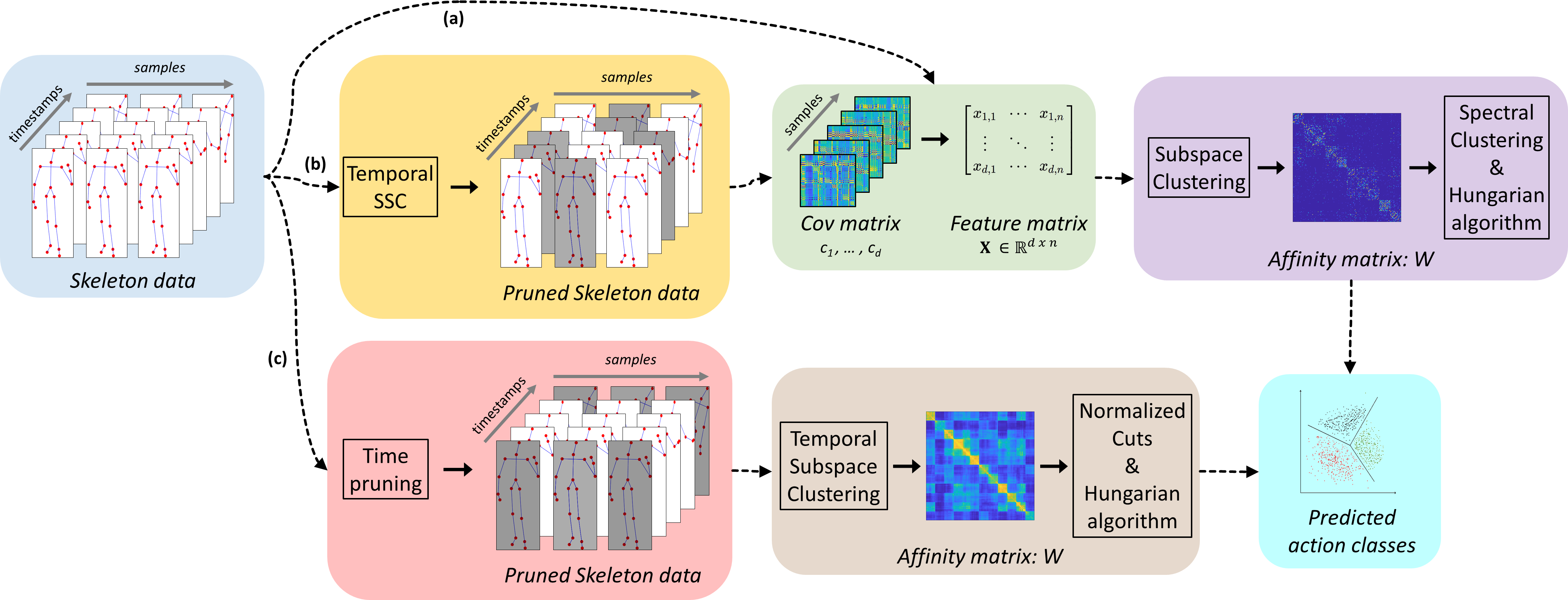}
\caption{Pipeline of the proposed unsupervised methods for HAR: (a) A covariance descriptor %(Eq. \eqref{eq:cov}) 
is applied to each sample. Given the obtained covariance matrix is square and symmetrical, we take only the upper (can be also lower) triangular part including the diagonal and flatten it. This results in a new matrix ($X$) having size $samples \times features$. Following that, any subspace clustering technique can be applied to obtain an affinity graph matrix $\mathbf{W}$. Then, spectral clustering is applied using $\mathbf{W}$ to obtain cluster labels and the Hungarian algorithm finds the matching between the cluster labels (predicted action classes) and the ground-truth labels. %(Section~\ref{sec:SSC})
(b) The skeletal data of each sample is temporally pruned using temporalSSC %(Section~\ref{sec:timeSSC}), 
and then the pruned data is processed as in (a).
(c) Each sample is pruned
by using various strategies. 
Afterwards, temporal subspace clustering is applied to obtain an affinity graph matrix $\mathbf{W}$. The normalized cuts is applied to obtain cluster labels and the Hungarian algorithm matches the cluster labels with the ground-truth labels.}
\label{fig:pipeline}
\end{figure*}

In this Section, we present our computational pipeline which is based on covariance representations and timestamps pruning. In order to properly ablate on the relative importance of them, we will consider the following computational variants of the pipeline:
%three unsupervised approaches to perform HAR. \textcolor{red}{C:at this point saying 3 different approaches is confusing because many previous time, we referred ''proposed pipeline'', which should be unified. Instead, we can say given the proposed pipeline \ref{fig:pipeline}, we apply the following experimental settings.}
\begin{itemize}
    \item \emph{Section~\ref{sec:SSC} and Figure \ref{fig:pipeline}(a).} We apply covariance encoding as the descriptor, whose result is given as an input to a subspace clustering method that is based on the self-expressiveness property of data. % (Section~\ref{sec:SSC}). This method is illustrated in F.
    \item \textit{Section~\ref{sec:timeSSC} and Figure \ref{fig:pipeline}(b).} We apply the proposed temporal pruning approach (namely \emph{temporalSSC}) as a pre-processing stage while the rest of the pipeline follows the previous setting.%for the following clustering stage based on the first proposal. %This method is integrated to the clustering methods. 
    %This approach shown in Figure \ref{fig:pipeline}(b).
    \item \textit{Section~\ref{sec:TSC} and Figure \ref{fig:pipeline}(c).} %\textcolor{red}{to add after figure 1 is confirmed: We apply covariance encoding as the descriptor, whose result is given as an input to...} 
    We use the Temporal Subspace Clustering to show the effectiveness of a dictionary-based subspace clustering for temporal series of data when applying temporal regularization on top of the (optional) encoding through covariance. %This final method is shown in Figure \ref{fig:pipeline}(c).%, resulting in good performance and outperforms even the supervised state-of-the-art methods on some datasets.
\end{itemize} 
\subsection{Subspace clustering methods based on self-expressiveness property and covariance representatio}\label{sec:SSC}

%\begin{figure*}[t!]\label{fig:SSC}\centering
%\includegraphics[width=\textwidth]{ssc}
%\caption{Pipeline of Subspace Clustering using self-expressiveness property of data: a covariance descriptor is applied to each sample of the original skeleton-based datasets.
%Given the obtained covariance matrix is square and symmetrical, we take only the upper (can be also lower) part of it from the diagonal and flatten it, which results in a new matrix having size $samples \times features$.
%Following that, different subspace clustering techniques are applied to obtain an affinity graph matrix; $W$.
%Spectral clustering is applied using $W$ to obtain cluster label and the Hungarian algorithm finds the matching and accuracy of cluster labels and ground-truth labels.}
%\end{figure*}

%\textcolor{red}{C:given that we wrote Figure 1(a).... do we still keep the list given above?-- a bit repetition. On the other hand such an indication Figure 1(a) was not done for other subsections e.g., Section 3-B.}
%Figure~\ref{fig:pipeline}(a) depicts the pipeline of this methodology, which has two main components: 
The usage of a covariance representation as the data encoder and the subspace clustering for solving HAR can be described as follows. %Each component is presented in this Section.

{\bf Data encoding through covariance representation.} Through either a motion capture system or a depth sensor, an action is represented as the collection in time of $K$ joints 3D positions $\mathbf{p}_1(t),\dots,\mathbf{p}_K(t)$.
By using $\mathbf{p}(t)$ to denote the column vectorization of all such 3D positions for a fixed timestamp, we represent an action sequence as the covariance matrix
\begin{equation}\label{eq:cov}
\boldsymbol{\Lambda} = \frac{1}{T} \sum_t (\mathbf{p}(t) - \boldsymbol{\mu})(\mathbf{p}(t) - \boldsymbol{\mu})^\top,\vspace{-5pt}
\end{equation}
where $T$ denotes the number of timestamps and $\boldsymbol{\mu}$ is the temporal average of $\mathbf{p}(t)$. 

We then vectorize the covariance matrix through a flattening operation which exploit the property of $\boldsymbol{\Lambda}$ in being symmetrical.
That is, $\boldsymbol{\Lambda} = \boldsymbol{\Lambda}^\top$.
Therefore, when flattening, we extract the diagonal elements of $\boldsymbol{\Lambda}$ (which are $\Lambda_{ii}$) and the upper-triangular ones (that is, $\Lambda_{ij}, j > i$).
The lower triangular part can be ignored since it is equal to upper triangular one.
Such flattening operation casts the $3K \times 3K$ matrix $\boldsymbol{\Lambda}$ into a $3K \cdot (3K - 1) / 2$ column vector. The flattened covariance representation is used as one data point, which then given to the subspace clustering algorithm as the input.

{\bf Subspace Clustering.} Let us consider a collection of $D$-dimensional data-points $\mathbf{x}_1,\dots,\mathbf{x}_N$.
Subspace clustering \cite{vidal2011subspace} attempts to cluster $\mathbf{x}_1,\dots,\mathbf{x}_N$ into groups (termed \emph{subspaces}) which share common geometrical relationships as the well-known \emph{self-expressiveness property}.
The problem can be formalised as finding a $N \times N$ matrix $\mathbf{C}$ of coefficients such that
\begin{equation}\label{eq:self-express}
    \mathbf{X} = \mathbf{X}\mathbf{C}~~\mathrm{subject}~\mathrm{to}~~{\rm diag}(\mathbf{C}) = 0,
    \vspace{-8pt}
\end{equation} 
where $\mathbf{X}$ is the $D \times N$ matrix, which stacks by columns the data points $\mathbf{x}_j$. The constraint ${\rm diag}(\mathbf{C}) = 0$ avoids the trivial solution corresponding to $\mathbf{C}$ being the identity matrix.
%(\textcolor{red}{FIXME: Mention the role of the C coefficients for the clustering problem!})
Ultimately, the geometrical relationship that we are interested in modelling is a linear relationship in which each data-point can be described as a linear combination. As a consequence of that, the subspaces are linear in turn.
The constraint ${\rm diag}(\mathbf{C}) = 0$ is fundamental to avoid the trivial (and useless) solution $\mathbf{x}_j = \mathbf{x}_j$. Specifically, the self-expressiveness property \eqref{eq:self-express} attempts to estimate each data points as a linear combination of \emph{different data points}. This allows to capture the geometrical inter-dependencies among the data points themselves.

An important aspect regarding subspace clustering is the way the matrix $\mathbf{C}$ is obtained. A number of works proposed to solve this problem through optimization \cite{lu2012robust,elhamifar2013sparse,ji2014efficient,you2016scalable,you2016oracle,ji2017deep} and different strategies have been adopted to constraint the solution.
In subspace segmentation via Least Squares Regression \textbf{(SS-LSR)} \cite{lu2012robust}, a Frobenius norm is introduced to promote a $L^2$ penalty, obtaining
\begin{equation}\label{eq:SS-LSR}
    \min \|\mathbf{C}\|_F~~\mathrm{subject}~\mathrm{to}~~\mathbf{X} = \mathbf{X}\mathbf{C},{\rm diag}(\mathbf{C}) = 0.    
\end{equation}

Another popular manner of constraining the coefficient matrix $\mathbf{C}$ is to impose sparsity \cite{elhamifar2013sparse,you2016oracle,ji2017deep}.
As in the Sparse Subspace clustering via Alternating Direction Method of Multipliers \textbf{(SSC-ADMM)} \cite{elhamifar2013sparse}, the problem formulation is framed as
\begin{equation}\label{eq:SSC-ADMM}
    \min \|\mathbf{C}\|_1~~\mathrm{subject}~\mathrm{to}~~\mathbf{X} = \mathbf{X}\mathbf{C},{\rm diag}(\mathbf{C}) = 0,    
\end{equation}
while using the alternating direction method of multipliers (ADMM) algorithm to foster convergence by solving a stack of easier sub-problems. 
As an alternative to ADMM, Sparse Subspace Clustering by Orthogonal Matching Pursuit \textbf{{(SSC-OMP)}} \cite{you2016scalable} approaches a similar problem with a different optimization technique. 

The previous formalism in Eq. (\ref{eq:SSC-ADMM}) was extended in the Deep Subspace Clustering Networks \textbf{(DSC-Nets)} \cite{ji2017deep} by having the hidden layer of an autoencoder implementing either equation \eqref{eq:SS-LSR} or equation \eqref{eq:SSC-ADMM}.
The Elastic Net \textbf{(EnSC)} \cite{you2016oracle} approach uses a convex combination of $L^2$ and $L^1$ constraint on $\mathbf{C}$ to increase performance, while also boosting the scalability due to the usage of oracle sets to better pre-condition the solution.
Dense subspace clustering \textbf{(EDSC)} \cite{ji2014efficient} approaches the problem by attempting to apply the self-expressiveness loss on a dictionary which is used to describe the data, while also taking into account outliers. 

Once the matrix of coefficient $\mathbf{C}$ is found, an affinity graph matrix $\mathbf{W}$ is built by setting the weights on the edges between the nodes through $\mathbf{W}=\mathbf{C}+\mathbf{C}^\top$.
Spectral clustering is later applied to $\mathbf{W}$ to obtain the clustering labels, by assigning each of the $N$ datapoint $\mathbf{x}_j$ into its corresponding subspace.
The final step is therefore apply Hungarian algorithm to compare and map subspace labels into actual class labels \cite{vidal2011subspace}.

%%%%%%%%%%%%%%%%%%%%%%%%%%%%%%%%%%%%%%%%%%%%%%%%%%
\subsection{Temporal pruning via Sparse Subspace Clustering (temporalSSC)}\label{sec:timeSSC}

In addition to utilize subspace clustering as an unsupervised learning method to perform action recognition, in this paper, we also exploit such family of techniques to solve another task: temporal pruning.
That refers to utilizing subspace clustering on the raw joint coordinates $\mathbf{p}(t)$.
Here, different from the previous section, each data point to be clustered is not an action sequence, but a single data point of an action (Figure~\ref{fig:pipeline}(b)). 
%\textcolor{red}{C: [sometimes in the text we refer an action sample as data point, so here saying not an action sequence but a data point is a bit confusing.]}
In other words, rather than applying subspace clustering to group action sequences, we exploit subspace clustering to the group skeletal poses at a given timestamp.
Our assumption is that the processed skeleton data might contain similar or even redundant poses over time. To address this, %In this manner, 
we apply temporal pruning, which potentially captures the similarities over time with respect to %and group timestamps accordingly to 
the kinematic execution. %{behind that is that utilizing the complete data, which some of them can be noisy, very similar to each other or redundant) our purpose is to find a better representational grouping of these given timestamps.}
%\textcolor{red}{FIXME: we should convey here a general idea why this is useful or at least which is the underlying assumption we are trying to verify...}

A relevant parameter for temporal pruning is the number of subspaces $\phi$, which corresponds to the length of the new pruned skeleton data, which was set based on the following strategies.%: once found, \textcolor{red}{FIXME: what?} the indexes of the subspaces help in performing a temporal pruning of the given sequence.
%In fact, 
%\textcolor{red}{FIXME: rephrase here, this description is not clear and does not flow well}.
%\textcolor{orange}{We heuristically determined $\phi$ by using different strategies:}
\subsubsection{min $\phi$} the temporal length of the entire dataset is fixed to be equal to the shortest time duration across all the sequences in the skeletal dataset, this is done by using the random permutation of each sample timestamps.
\subsubsection{min temporalSSC} subspace clustering method SSC\_ADMM is used to get $\phi$ equal to the shortest time duration across all the sequences in the skeletal dataset.
\subsubsection{percentage temporalSSC} the temporal length of each sample of the dataset is determined by selecting a percentage value for $\phi$ (in our experiments we chose to keep the 75\%, 50\% or 25\% of the sample temporal length) and applying temporalSSC.
\subsubsection{threshold temporalSSC} the temporal length of each sample of the dataset is determined by selecting a percentage value for $\phi$ (in our experiments we chose to keep the 75\%, 50\% or 25\% of the sample temporal length), which is used as a threshold value for temporalSSC. If a certain sample of the dataset has a temporal length superior to $\phi$, temporalSSC is therefore applied to match this threshold value. 

Once $\phi$ is fixed according to one of the previous strategy, we can now retrieve all the timestamps $t_1,\dots,t_s,\dots$ assigned to a given subspace. Afterwards, we average the corresponding skeletal positions $\mathbf{p}(t_1),\dots,\mathbf{p}(t_s),\dots$. The so-obtained average skeletal position is adopted to replace the original one and the procedure is iterated across all the different subspaces. For the sake of clarity, let us exemplify the procedure in a particular case. For instance, lets assume that the number of subspaces is set to be $\phi = 2$ and the original action sequence has 5 timestamps to which are associated the following body poses $[\mathbf{p}_1,\mathbf{p}_2,\mathbf{p}_3,\mathbf{p}_4,\mathbf{p}_5]$. Once temporalSSC is runned on top of the sequence $[\mathbf{p}_1,\mathbf{p}_2,\mathbf{p}_3,\mathbf{p}_4,\mathbf{p}_5]$, let assume that the corresponding output is $[1,1,2,1,2]$. So, temporalSSC is grouping $\mathbf{p}_1,\mathbf{p}_2$ and $\mathbf{p}_4$ in a subspaces and $\mathbf{p}_3,\mathbf{p}_5$ in another one. Then, we  define the pruned action sequence as $[\mathbf{p}'_1,\mathbf{p}'_2]$, where $\mathbf{p}'_1 = \frac{1}{3}(\mathbf{p}_1+\mathbf{p}_2+\mathbf{p}_4)$ and $\mathbf{p}'_2 = \dfrac{1}{2}(\mathbf{p}'_3+\mathbf{p}'_5)$.

%\textcolor{red}{C: At which stage this is done, it pseudo labels this is applied on, it is not clear.--> We consider all timestamps $t_1,\dots,t_s,\dots$ assigned to a given subspace.
%We then average the corresponding skeletal positions $\mathbf{p}(t_1),\dots,\mathbf{p}(t_s),\dots$.
%We then replace the original skeletal positions with the averaged positions obtained in the method we just described.}

Once the temporal pruning is performed, the covariance representation is applied to the new data and subspace clustering is adopted as in Section~\ref{sec:SSC}.
%Optionally\footnote{Referring to Section \ref{sec:datasets}, we performed the referred data augmentation only for Florence3D and UTKinect datasets, due to the extremely short temporal duration of the acquired sequences.}, before performing the temporal pruning, we exploited \cite{zanfir2013moving} as a data augmentation strategy to pair 3D positions with velocities and accelerations.

%%%%%%%%%%%%%%%%%%%%%%%%%%%%%%%%%%%%%%%%%%%%%%%%%%
\subsection{Temporal Subspace Clustering based on dictionary and temporal Laplacian Regularization}\label{sec:TSC}

Even though subspace clustering methods explained in Section~\ref{sec:SSC} build the affinity matrix $\mathbf{W}$ by exploiting the self-expressiveness property of data, they do not explicitly take into account the temporal dimension of time-series data while building the model adopted for HAR.
%. This approach surely gives promising results (as reported in Section~\ref{sec:res_SSC} and \ref{sec:res_timeSSC} with the addition of timeSSC) but optimal \textcolor{red}{FIXME: optimal in which sense?} performance on encoding time-series data (in the case of this paper, HAR on skeleton-joints datasets) could be hindered \textcolor{red}{FIXME: not clear why, explain}.
%In other words, by encoding the temporal dimension of this kind of data into covariance descriptors, the correlation between adjacent temporal sequences of a given data sample could not be efficiently encoded.
As a solution, temporal regularization was proposed by Temporal Subspace Clustering \textbf{(TSC)} \cite{li2015temporal}. Precisely, given a dictionary $\mathbf{D} \in \mathbb{R}^{d \times r}$ and a coding matrix $\mathbf{Z} \in \mathbb{R}^{r \times n}$, a collection of data points $\mathbf{X} \in \mathbb{R}^{d \times n}$ can be approximately represented as
\begin{equation}
    \mathbf{X} \approx \mathbf{DZ},
\end{equation}
where each data point is encoded using a Least Squares regression, and a temporal Laplacian regularization $L(Z)$ function encourages the encoding of the sequential relationships in time-series data. This can done by minimising
\begin{equation}
\begin{gathered}
\min_{\mathbf{Z},\mathbf{D}}~~\|\mathbf{X} - \mathbf{DZ}\|^2_F~+~\lambda_1\|\mathbf{Z}\|^2_F~+~\lambda_2L(\mathbf{Z}),\\
subject~to~~\mathbf{Z}\geq0,~\mathbf{D}\geq0,
\end{gathered}
\end{equation}
by using the ADMM algorithm to encourage convergence by solving a stack of easier sub-problems. Different from Section \ref{sec:SSC}, % shows how subspace clustering methods based on the self-expressiveness property of data build the affinity graph matrix $\mathbf{W}$ by using $\mathbf{W} = \mathbf{C} + \mathbf{C}^T$, however the intrinsic relationships of within-cluster samples could not be efficiently exploited. Therefore, in \cite{li2015temporal} 
the affinity graph matrix $\mathbf{W}$ is given by the coding matrix $\mathbf{Z}$ by using $\mathbf{W}(i,j) = \frac{z_i^\top z_j}{||z_i||_2 ||z_j||_2}$, since the within-cluster samples (for example the sequential neighbors of a time-series datapoint) are always highly correlated to each other \cite{li2013low, li2014learning}. As final steps of the pipeline, the standard Normalized Cuts \cite{shi2000normalized} and Hungarian algorithms determine the clustering labels necessary for evaluation against the ground-truth.

%\textcolor{red}{C: The 'Observation' part should be moved to section 5-C as in figure 1 we kept covariance.} %J: no, I do not like it. The method should be self-contained in Section III. It's weird to me to have some of the pipeline in Section V
%{\bf Observation.} 
In Sections~\ref{sec:SSC} and \ref{sec:timeSSC}, a (flattened) covariance representation was adopted to encode the actions' kinematics. Computationally, this operation was able to cast an action sequence with a variable temporal duration into a fixed-size embedding which was passed in input to subspace clustering methods based on the self-expressiveness property. Here, differently, TSC leverages a dictionary learning framework which, together with the temporal regularization, should be effective in capturing the temporal variability of the data. To understand to which extent this is true, we would like to \emph{intentionally} get rid of covariance representations within our computational pipeline in order to separately evaluate this two alternative strategies of handling the temporal dimensions of the data.
%to which extent the temporal regularization of TSC is effective for the sake of HAR. 
%The shape of our skeleton-joints dataset for HAR does is \textcolor{red}{FIXME: this is out of the blue...why do we need this? we already formalised the math...can we use a more rigorous formalisation?}:\\
%$[samples ~ \times ~ xyz\_coordinates ~ \times ~ joints ~ \times ~ timeframes]$\\
%where each distinct sample got a different length of $timeframes$.
%For the methodologies of Section~\ref{sec:SSC} and \ref{sec:timeSSC} this is not an issue because, by using a covariance descriptor \eqref{eq:cov}, the data matrix shape becomes $[samples ~ \times ~ flattened ~ cov ~ features]$ (thus encoding the temporal frames into a fixed length) \textcolor{red}{flattening means $vec$ operator and $vech$ for symmetric matrices, we are loosing all the formalism here, anyway still not very convinced about this paragraph...}.
%So, in order to use the TSC approach without a covariance descriptor as input, we designed different approaches to set 

TSC approach is combined with the following pruning strategies such that a constant temporal length $\phi$ for all the dataset in use is set as:
\subsubsection{TSC min} the temporal length $\phi$ of the entire dataset is fixed to be equal to the shortest time duration across all the sequences in the skeletal dataset, this is done by using the random permutation of each timeframe.
\subsubsection{TSC max} the opposite process of \emph{TSC min}. For each instance, its timeframes are replicated until the temporal length $\phi$ is equal to the longest time duration across all the sequences in the skeletal dataset.
\subsubsection{temporalSC + TSC} spectral clustering is used to get $\phi$ equal to the shortest time duration across all the sequences in the skeletal dataset.
\subsubsection{temporalKm + TSC} k-means clustering is used to get $\phi$ equal to the shortest time duration across all the sequences in the skeletal dataset.

%After performing a concatenation of $[xyz\_coordinates ~ \times ~ joints ~ \times ~ fixed ~ timeframes]$ \textcolor{red}{FIXME:better way to write this?}, the new data matrix has the shape $[samples ~ \times ~ features]$, which is given to the TSC algorithm \cite{li2015temporal} as an input to obtain the affinity matrix $W$.

%Following that, Normalized Cuts and Hungarian algorithm are applied to compare and map subspace labels to the actual class labels This is illustrated in Figure~\ref{fig:pipeline}(c).

%%%%%%%%%%%%%%%%%%%%%%%%%%%%%%%%%%%%%%%%%%%%%%%%%%
\section{3D action recognition dataset}\label{sec:datasets}
There exists a consistent variability in every HAR dataset due to the length in the performed actions and their complexity, the number of action classes and the technology that was used to capturing them. Prior to experimental analysis, a pre-processing step is performed \cite{Camps:ACCV14, Camps:CVPR16, Vemulapalli:CVPR14, Vemulapalli:CVPR16, ECCV16, Cavazza:ICPR16, Liu:ECCV16} in order to fix one root joint located at the hip center, and compute the relative differences of all other $J-1$ 3D joint positions. This pre-processing is performed at any timestamps $t=1,\dots,T$ to obtain a 3($J-1$)-dimensional (column) vector $p(t)$ of the relative displacements. We used the following dataset for our experimental analysis. 
\newline
\textbf{Florence3D (F3D) \cite{seidenari2013recognizing}}: a 9-class action dataset (\emph{answer phone, bow, clap, drink, read watch, sit down, stand up, tight lace, wave}) captured using a Microsoft Kinect camera. The actions were performed for two/three times by 10 subjects, resulting in 215 data samples.
\newline
\textbf{UTKinect-Action3D (UTK) \cite{xia2012view}}: a 10-class action dataset (\emph{carry, clap hands, pick up, pull, push, sit down, stand up, throw, walk, wave hands}) captured using a single stationary Microsoft Kinect camera. Each action was performed for two times by 10 subjects, resulting in 199 data samples.
\newline
\textbf{MSR 3D Action Pairs (MSRP) \cite{oreifej2013hon4d}}: includes 12 actions in pairs (\emph{pick up box, put down box, lift box, place box, push chair, pull chair, wear hat, take off hat, put on backpack, take off backpack, stick poster, remove poster}). Each pair has similar features but their relation in terms of motion and shape is different. The actions were performed for three times by 10 subjects, resulting in 353 activity samples.
\newline
\textbf{MSR Action 3D (MSRA) \cite{li2010action}}: a 20-class action dataset (\emph{bend, draw circle, draw tick, draw x, forward kick, forward punch, golf swing, hand catch, hand clap, hammer, high arm wave, high throw, horizontal arm wave, jogging, pick up and throw, sideboxing, side kick, tennis serve, tennis swing, two-handwave}) captured by a depth-camera. Each action was performed for three times by 10 subjects, resulting in 557 data samples.
\newline
\textbf{Gaming 3D (G3D) \cite{bloom2016hierarchical}}: a 20-class gaming actions dataset (\emph{aim and fire gun, clap, climb, crouch, defend, flap, golf swing, jump, kick left, kick right, punch left, punch right, run, steer a car, tennis swing backhand, tennis swing forehand, tennis serve, throw bowling ball, wave, walk}) captured using a Kinect camera. The actions were repeated for seven times by 10 subjects, resulting in 663 activity samples.
\newline
\textbf{HDM05 \cite{cg-2007-2}}: due to class imbalance of the original dataset, we select 14 classes (\textbf{HDM-05-14}, \emph{clap above head, deposit floor, elbow to knee, grab high, hop both legs, jog, kick forward, lie down on floor, rotate both arms backward, sit down chair, sneak, squat, stand up, throw basketball}, following the protocol of \cite{Wang:ICCV15, Cavazza:ICPR16}), and 65 classes (\textbf{HDM-05-65}, following the protocol of \cite{bellolui} by grouping together similar actions). The sequences were captured using VICON cameras, resulting in 686 data samples for the former and 2343 data samples for the latter.
\newline
\textbf{MSRC-Kinect12 (MSRC) \cite{fothergill2012instructing}}: a 12-class gesturing dataset, grouped into iconic and metaphoric gestures (\emph{beat both, bow, change weapon, duck, goggles, had enough, kick, lift outstretched arms, push right, shoot, throw, wind it up}). Highly corrupted actions were removed following the protocol as in \cite{egizi}, resulting in 5881 data samples.

%%%%%%%%%%%%%%%%%%%%%%%%%%%%%%%%%%%%%%%%%%%%%%%%%%
\section{Experimental Analysis}\label{sec:experiments}

In this section we evaluate the methodologies presented in Section~\ref{sec:method}, applied on the dataset reported in Section~\ref{sec:datasets}:
%The reader can follow in Figure~\ref{fig:pipeline} a graphical description of the three different pipelines: % of this experimental section: namely,  to build the affinity matrix $W$, namely:\\
The Subspace Clustering methods based on self-expressiveness property of data (Section~\ref{sec:SSC}), 
%where the skeleton joints datasets are first encoded by using a covariance descriptor  and the following flattened input matrix is applied to the different subspace clustering methods taken in examination (Section~\ref{sec:SSC}).\\
 the Temporal pruning via Sparse Subspace Clustering (temporalSSC) (Section~\ref{sec:timeSSC}) and the
%: where the skeleton joints datasets are first pruned along their temporal dimension (Section~\ref{sec:timeSSC}) by using the Sparse Subspace Clustering via Alternating Direction Method of Multipliers (SSC) algorithm \cite{elhamifar2013sparse}, followed by the application of a covariance descriptor and the flattened input matrix is applied to SSC.\\
Temporal Subspace Clustering based on dictionary and temporal Laplacian Regularization (Section~\ref{sec:TSC}).\\
%: where we uniform the temporal length of the different skeleton joints datasets (Section~\ref{sec:TSC}) and the flattened input matrix is applied to the Temporal Subspace Clustering (TSC) \cite{li2015temporal}.
%\textcolor{red}{FIXME: better to use understandable methods acronyms instead of A, B, C} %J: I directly removed the letters.

\noindent {\bf Error metrics and performance evaluation.} To monitor the performance in HAR, we will take advantage of classification accuracy defined as  
\begin{equation}
    ACC (\%) = \left(1 - \frac{\# ~ of ~ misclassified ~ labels}{\# ~ of ~ total ~ labels}\right) \times 100
\end{equation}
and expressed as a percentage. As explained in Section \ref{sec:method}, the clustering labels are obtained through either spectral clustering \cite{pedregosa2011scikit} or Normalized Cut \cite{shi2000normalized}. Finally, the Hungarian algorithm \cite{vidal2011subspace} maps cluster labels into the ground-truth ones.

%%%%%%%%%%%%%%%%%%%%%%%%%%%%%%%%%%%%%%%%%%%%%%%%%%
\subsection{Subspace Clustering methods based on self-expressiveness property of data}\label{sec:res_SSC}

\begin{table}[!t]\centering
\caption{Clustering accuracy (\%) of subspace clustering methods as well as k-means (Km) and spectral clustering (Sc). AVG and STD stand for the average and standard deviation of results at each column. The best performance for each dataset emphasized in bold.}
\label{table:SSC}
\resizebox{\columnwidth}{!}{
\begin{tabular}{|l|c|c|c|c|c|c|c|c|} \hline
Dataset   & Km    & Sc    & EDSC           & OMP   & DSCN           & LSR            & SSC   & EnSC           \\ \hline
F3D       & 45,58 & 66,05 & 54,42          & 61,40 & 57,02          & 60,47          & 69,12 & \textbf{70,23} \\
UTK       & 34,67 & 66,83 & 52,71          & 58,79 & 69,35          & 57,79          & 73,97 & \textbf{78,90} \\
MSRP      & 42,78 & 52,69 & \textbf{51,90} & 50,14 & 49,26          & 47,31          & 49,60 & 49,86          \\
MSRA      & 41,11 & 65,17 & 52,69          & 43,99 & 59,91          & 54,40          & 57,27 & \textbf{62,84} \\
G3D       & 31,22 & 64,71 & 44,48          & 45,70 & 62,59          & 64,25          & 65,16 & \textbf{72,25} \\
HDM-05-14 & 32,36 & 53,35 & 52,42          & 47,67 & \textbf{56,27} & 51,60          & 49,13 & \textbf{56,00} \\
HDM-05-65 & 31,41 & 44,46 & \textbf{44,43} & 36,07 & 30,95          & 42,98          & 35,98 & 42,38          \\
MSRC      & 61,54 & 84,34 & 81,30          & 51,20 & 71,35          & \textbf{87,04} & 62,27 & 83,27          \\ \hline
AVG       & 40,08 & 62,22 & 54,29          & 49,37 & 57,09          & 58,23          & 57,81 & 64,46          \\
STD       & 09,63 & 11,28 & 10,82          & 07,58 & 11,92          & 12,66          & 11,63 & 13,38          \\ \hline
\end{tabular}}
\end{table}

%\textcolor{red}{Make clear here if you evaluate our methods A,B, C or just the baseline, say clearly in the beginning which is the purpose of this section!}
In this section we experimentally validate the computational method presented in Section~\ref{sec:SSC} and visualised in Figure \ref{fig:pipeline}(a): In order to exploit the self-expressiveness property of data and to encode their temporal information, we implement the covariance descriptor to encode the raw data. % into a new input matrix of shape $[samples ~ \times ~ flattened ~ cov ~ features]$ \textcolor{red}{FIXME: again...}.

Following that, we used the state-of-the-art subspace clustering methods that are based on the self-expressiveness property of the data to obtain the affinity matrix $\mathbf{W}$. These methods are: %Efficient Dense Subspace Clustering (
EDSC \cite{ji2014efficient}, %Sparse Subspace Clustering by Orthogonal Matching Pursuit (
OMP \cite{you2016scalable}, %Deep Subspace Clustering Networks 
DSCN \cite{ji2017deep}, %Subspace Segmentation via Least Squares Regression (
LSR \cite{lu2012robust}, %Sparse Subspace Clustering via Alternating Direction Method of Multipliers (
SSC \cite{elhamifar2013sparse}, %Elastic net Subspace Clustering (
EnSC \cite{you2016oracle} which are described in Section \ref{sec:SSC}. %\textcolor{red}{FIXME: no need to repeat the names if we did already...use just acronyms  with citations} %J: ok

Once the coefficient matrix $\mathbf{C}$ and the affinity graph matrix $\mathbf{W}$ were found, spectral clustering and Hungarian algorithm were applied to map the subspace label with the actual class labels \cite{vidal2011subspace} as illustrated in Figure \ref{fig:pipeline}(a).

Additionally, as a baseline method, we considered two of the most popular clustering method: K-means clustering \textbf{(Km)} and spectral clustering \textbf{(Sc)} \cite{pedregosa2011scikit} and all the corresponding results are reported in Table~\ref{table:SSC}. The overall best performing method is Elastic net Subspace Clustering (EnSC) \cite{you2016oracle}, which ranked highest for five of the nine datasets. For three of these five, i.e., UTKinect, MSRAction3D, and G3D datasets, EnSC's performance is approximately 5\% better than the second best performing method.

%%%%%%%%%%%%%%%%%%%%%%%%%%%%%%%%%%%%%%%%%%%%%%%%%%
\subsection{Temporal pruning via Sparse Subspace Clustering (temporalSSC)}\label{sec:res_timeSSC}

\begin{table}[!t]\centering
\caption{Clustering accuracy (\%) of temporalSSC combined with different strategies and when standard SSC applied for the final clustering. The first column shows the SSC's performances alone. AVG and STD stand for the average and standard deviation of results at each column. Best performance of each dataset emphasized in bold.}
\label{table:timeSSC}
\resizebox{\columnwidth}{!}{
\begin{tabular}{|l|c|c|c|c|c|c|c|}
\cline{3-5} \cline{7-7}
\multicolumn{2}{c|}{} &
\multirow{2}{*}{\begin{tabular}[c]{@{}c@{}}min\\$\phi$\end{tabular}} &
\multirow{2}{*}{\begin{tabular}[c]{@{}c@{}}min\\temporalSSC\end{tabular}} &
\multirow{2}{*}{\begin{tabular}[c]{@{}c@{}}percentage\\temporalSSC\end{tabular}} &
\multicolumn{1}{c|}{} &
\multirow{2}{*}{\begin{tabular}[c]{@{}c@{}}threshold\\temporalSSC\end{tabular}}\\
\cline{1-2} \cline{6-6} \cline{8-8}
Dataset   & SSC            &                &                &                & $\phi$ &                & $\phi$ \\ \hline
F3D       & \textbf{69,12} & 67,91          & 66,51          & 65,12          & 75\% & 68,84          & 50\% \\
UTK       & 73,97          & 64,82          & \textbf{80,90} & 68,34          & 25\% & 72,86          & 75\% \\
MSRP      & 49,60          & 48,88          & 47,88          & \textbf{50,42} & 25\% & 49,58          & 25\% \\
MSRA      & 57,27          & 59,61          & 57,09          & 62,66          & 25\% & \textbf{63,02} & 75\% \\
G3D       & 65,16          & 64,86          & 64,10          & 69,68          & 75\% & \textbf{71,49} & 75\% \\
HDM-05-14 & 49,13          & \textbf{63,12} & 59,04          & 59,33          & 25\% & 59,77          & 25\% \\
HDM-05-65 & 35,98          & 41,31          & \textbf{44,00} & 43,66          & 25\% & 41,53          & 50\% \\
MSRC      & 62,27          & \textbf{83.79} & 83,62          & 83,41          & 75\% & 83,14          & 75\% \\ \hline
AVG       & 57,81          & 61,79          & 62,89          & 62,83          &      & 63,78          &      \\
STD       & 11,63          & 11,90          & 13,23          & 11,40          &      & 12,53          &      \\ \hline
\end{tabular}}
\end{table}

This pipeline is similar to \emph{(A)} but we applied to raw data, before the encoding of the covariance descriptor, different pruning strategies for the temporal dimension of data by using SSC (see Figure~\ref{fig:pipeline}(b)). For the subspace clustering implementation, we decided to use SSC for its computational efficiency and rapid convergence time.

Table~\ref{table:timeSSC} reports the clustering accuracy of different temporalSSC strategies, along with SSC results of Table~\ref{table:SSC} \cite{elhamifar2013sparse} as a baseline comparison.
Results of \emph{percentage temporalSSC} and \emph{threshold temporalSSC} are related to the best accuracy along the different percentage values of $\phi$ (i.e. 75\%, 50\% and 25\%).
Only with the exception of F3D (due to its original low dimensionality of the dataset and the extreme pruning of timestamps), the results show that applying temporalSSC overall contributes positively to the clustering performance of SSC \cite{elhamifar2013sparse}: The performance improvement is up to an average 8\% among all dataset, where on MSRC (the biggest dataset available) the improvement goes up to 21\%.

\subsection{Temporal Subspace Clustering based on dictionary and temporal Laplacian Regularization}\label{sec:res_TSC}

\begin{table*}\centering
\caption{Clustering accuracy (\%) of TSC combined with different strategies of uniforming temporal dimension of each dataset. The supervised state-of-the-art (s.o.t.a) results are also given. AVG and STD stand for the average and standard deviation of results at each column. Best unsupervised performance of each dataset emphasized in bold.}
\label{table:TSC}
%\resizebox{\textwidth}{!}{
\begin{tabular}{|l|c|c|c|c|c|c|c|c|c|}
\hline
\multirow{2}{*}{\begin{tabular}[c]{@{}c@{}}Dataset\end{tabular}} &
\multirow{2}{*}{\begin{tabular}[c]{@{}c@{}}TSCmin\end{tabular}} & \multirow{2}{*}{\begin{tabular}[c]{@{}c@{}}cov\\TSCmin\end{tabular}} &
\multirow{2}{*}{\begin{tabular}[c]{@{}c@{}}TSCmax\end{tabular}} & \multirow{2}{*}{\begin{tabular}[c]{@{}c@{}}cov\\TSCmax\end{tabular}} &
\multirow{2}{*}{\begin{tabular}[c]{@{}c@{}}temporalSC\\+ TSC\end{tabular}} & \multirow{2}{*}{\begin{tabular}[c]{@{}c@{}}temporalSC\\+ TSC cov\end{tabular}} &
\multirow{2}{*}{\begin{tabular}[c]{@{}c@{}}temporalKm\\+ TSC\end{tabular}} & \multirow{2}{*}{\begin{tabular}[c]{@{}c@{}}temporalKm\\+ TSC cov\end{tabular}} &
\multirow{2}{*}{\begin{tabular}[c]{@{}c@{}}supervised\\s.o.t.a.\end{tabular}}\\
          &                &             &                &             &                &                  &                &                  &                                   \\ \hline
F3D       & 84,65          & 81,40       & 94,88          & 81,86       & \textbf{95,81} & 88,84            & 87,91          & 87,44            & 99,07  \cite{li2018spatio}        \\
UTK       & 93,97          & 96,98       & \textbf{99,50} & 92,96       & 96,98          & 96,98            & 93,47          & 83,92            & 100,00 \cite{zhang2016efficient}  \\
MSRP      & 93,48          & 81,30       & \textbf{98,02} & 84,70       & 88,67          & 76,20            & 96,32          & 71,10            & 95,50  \cite{cavazza2019scalable} \\
MSRA      & 87,18          & 79,89       & 85,64          & 83,30       & 82,47          & 81,13            & \textbf{88,51} & 87,61            & 97,40  \cite{cavazza2019scalable} \\
G3D       & 88,99          & 90,20       & 85,07          & 92,61       & 90,20          & 92,46            & 88,84          & \textbf{92,91}   & 96,02  \cite{wang2018action}      \\
HDM-05-14 & \textbf{89,80} & 86,73       & 80,32          & 83,82       & 88,48          & 84,84            & 83,97          & 81,63            & 99,10  \cite{cavazza2019scalable} \\
HDM-05-65 & 70,51          & 83,57       & 75,97          & 85,62       & 72,13          & 84,64            & 68,42          & \textbf{86,00}   & 96,92  \cite{du2015hierarchical}  \\
MSRC      & 97,96          & 91,09       & \textbf{99,08} & 99,05       & 98,81          & 97,42            & 99,00          & 91,07            & 98,50  \cite{cavazza2019scalable} \\ \hline
AVG       & 88,32          & 86,40       & 89,81          & 87,99       & 89,19          & 87,81            & 88,31          & 85,21            &                                   \\
STD       & 7,79           & 5,59        & 8,62           & 5,72        & 8,18           & 7,05             & 8,80           & 6,31             &                                   \\ \hline
\end{tabular}
%}
\end{table*}

Table~\ref{table:TSC} reports the unsupervised clustering accuracy of the approach given in Section~\ref{sec:TSC} (as well as illustrated in Figure~\ref{fig:pipeline}(c)) is applied. %namely: \emph{TSCmin}, \emph{TSCmax}, \emph{timeSC + TSC}, \emph{timeKm + TSC}.
We also \emph{TSCmin}, \emph{TSCmax}, \emph{temporalSC + TSC}, \emph{temporalKm + TSC} with and without covariance descriptor.
%replicated these methods by implementing as well the covariance descriptor for a straight-on comparison (indicated by the suffix $\sim cov$).
The last column of that table reports the state-of-the-art performance obtained for each dataset. It is important to highlight that the corresponding state-of-the-art methods are all supervised while all other results given in that table are unsupervised.

The results show that the application of TSC gives the best overall accuracy among all techniques adopted in this paper, Table~\ref{table:TSC} demonstrates that the average of results of each implementation (column) is over over 85\% among all cases.
%
%Another point is the direct comparison between the methods applied by using raw data and the application of a covariance descriptor.
Except G3D and HDM-05-65 datasets, the average accuracy of each method without covariance (\emph{cov}) descriptor is approximately 2\% better than a method with \emph{cov} descriptor.
The comparisons between the temporal frames selection approaches show that in 5-out-of-8 datasets the pruning of data, therefore reduction of its temporal dimension, is beneficial to encode and represent this type datasets.
Whereas, for the datasets MSRP and MSRC, augmenting the data in temporal dimension leads to performance levels better than the state-of-the-art methods, which are all supervised.

% Note that the IEEE does not put floats in the very first column - or typically anywhere on the first page for that matter.
% Also, in-text middle ("here") positioning is typically not used, but it is allowed and encouraged for Computer Society conferences (but not Computer Society journals).
% Most IEEE journals/conferences use top floats exclusively.
% Note that, LaTeX2e, unlike IEEE journals/conferences, places footnotes above bottom floats.
% This can be corrected via the \fnbelowfloat command of the stfloats package.

%%%%%%%%%%%%%%%%%%%%%%%%%%%%%%%%%%%%%%%%%%%%%%%%%%
\section{Conclusion}\label{sec:conclusion}

Human Activity Recognition (HAR) is a challenging problem, which has been solved with different methodologies and the sharp majority of them apply a supervised learning paradigm. This paper particularly focuses on skeletal data analysis and, differently, embraces a fully unsupervised approach to tackle HAR. %This is unlike the state-of-the-art methods that typically apply supervised learning to solve for this complex problem.
In this study, we propose a novel clustering pipeline, which combines  covariance descriptors and subspace clustering applied to 1) temporally prune the input data and 2) group together similar activities based on their respective category. The aim of temporal pruning is to discriminate better the action sequences that are recognized with an unsupervised method.

The experimental analysis is validated on eight different dataset, which are different from each other in terms of action types, the number of action classes involved as well as the experimental protocol they were captured. Across such a wide variety of experimental benchmarks, our findings show that our proposed pipeline is superior to previous subspace clustering methods relying on the self-expressiveness property of data. 

%, but also sometimes performs as well as the state-of-the-art supervised approaches. %even favorably scores with respect to supervised approaches.
%In detail, the proposed pipeline performs better than the subspace clustering methods based on self-expressiveness property of data \textcolor{red}{FIXME: not clear this sentence}. 
Subspace clustering methods based on the self-expressiveness property can remarkably enhanced in performance by covariance representation to the point that other baseline methods are systematically outperformed.
On the other hand, temporal subspace clustering method that relies on dictionary learning and temporal Laplacian regularization combined within our pipeline results in remarkably good HAR performances: This demonstrates the benefits of pruning action sequences along the temporal dimension. Overall, the combination of our experimental findings enable a fully unsupervised pipeline for HAR to always reduce the gap with supervised approaches, while surprisingly outperforming them in some cases.
%such that in some datasets its performance is better than the state-of-the-art supervised methods.} 
%\textcolor{red}{FIXME: Rewrite conclusions, you can go more straight to the point, we even beat supervised methods!!!!!!}.

%\textcolor{modifica}{We provide an exhaustive experimental validation in which we compare the performance of a variety of subspace clustering methods applied to eight public recognition benchmark datasets for HAR. Overall, Elastic Net Subspace Clustering \cite{you2016oracle} }

\bibliographystyle{IEEEtran}
\bibliography{IEEEabrv,mybib,fonti}

\end{document}